\documentclass[journal]{IEEEtran}

\usepackage{makecell}
\usepackage{amsmath}
\usepackage{amssymb}
\usepackage{cite}
\usepackage{graphicx}
\usepackage{booktabs}
\usepackage{multirow}
\usepackage{subcaption}
\usepackage{hyperref}
\hypersetup{hidelinks}
\usepackage{cleveref}   % must come after hyperref

\begin{document}

\title{Edge Assisted Multi-Camera Vehicle Tracking Framework for Real-Time and Scalable Deployment}

\author{Yuqiang~Lin, Sam~Lockyer, Shucheng~Zhang, Florian~Stanek, Markus~Zarbock, Adrian~Evans, Wenbin~Li, Yinhai~Wang, and Nic~Zhang
\thanks{Y. Lin, S. Lockyer, A. Evans, W. Li, and N. Zhang are with the University of Bath, Bath, United Kingdom.}
\thanks{S. Zhang and Y. Wang are with the University of Washington, Seattle, WA, USA.}
\thanks{F. Stanek and M. Zarbock are with Starwit Technologies GmbH, Germany.}
\thanks{Corresponding author: N. Zhang.}
% TODO: add corresponding author's e-mail address in the \thanks above.
}

\maketitle

\begin{abstract}
Cameras are a core sensing modality in modern intelligent transportation systems (ITS), providing rich visual information about road-user activities. Multi-Camera Vehicle Tracking (MCVT) leverages this data to reconstruct vehicle trajectories across a camera network, enabling multiple applications. However, existing MCVT studies mainly emphasize tracking accuracy while overlooking real-time performance and scalability, both of which are critical for real-world deployment. To address this gap, we propose the Edge-Assisted, Scalable and Efficient-MCVT (EASE-MCVT) framework, explicitly designed for real-time throughput and scalable deployment. Specifically, EASE-MCVT follows an edge-server architecture. On the edge side, each camera stream is processed in real time through object detection, single-camera tracking, geo-mapping, and feature extraction, while only lightweight metadata, including vehicle locations and appearance features, are transmitted to the central server for cross-camera association. To improve both tracking performance and system efficiency, EASE-MCVT is further optimized from both algorithmic and system perspectives. On the algorithmic side, we introduce a dynamic workload scheme for tracklet-level feature extraction on edge devices, a server-side single-camera re-match module to reconnect fragmented tracklets, and a self-supervised camera link model that learns spatio-temporal constraints to accelerate and stabilize cross-camera association. On the system side, we integrate production-oriented data engineering components to standardize deployment and data exchange for large-scale operation. Experiments on RoundaboutHD and CityFlow demonstrate stable real-time throughput while maintaining competitive tracking accuracy, achieving 71.63\% IDF1 on RoundaboutHD and stable server-side processing with up to 32 emulated camera streams on the tested consumer-grade workstation.
\end{abstract}


\section{Introduction}
The rapid development of urbanization has posed new challenges for transportation authorities in maintaining efficient, safe, and sustainable traffic operations. As road networks become more complex and the number of vehicles continues to grow, conventional traffic management methods are often insufficient to meet increasing transportation demands \cite{BACKGROUND}. To address these challenges, leveraging data from existing traffic cameras has become a practical and cost-effective approach. These cameras continuously capture rich visual information about road environments, which can be processed using advanced computer vision technologies to provide data-driven insights for intelligent traffic management.

Among these technologies, multi-camera vehicle tracking (MCVT) has emerged as a particularly promising approach. The primary goal of MCVT is to associate vehicle trajectories across different cameras while maintaining consistent vehicle identities across different fields of view (FOVs) \cite{amosa2023review}. This capability enables the reconstruction of complete vehicle trajectories throughout a monitored area. When applied to multiple vehicles, MCVT can generate a comprehensive representation of traffic flow. The reconstructed traffic flow data can subsequently support various downstream applications, including traffic flow prediction, suspect vehicle tracking, and traffic anomaly detection. Collectively, these applications can serve as key components in building safer, smarter, and more efficient transportation systems.

Generally, the MCVT problem can be divided into two stages \cite{holla2025review}: \textit{(i) the intra-camera process}, which detects and tracks vehicles within the same camera view across consecutive frames; and \textit{(ii) the inter-camera process}, which associates vehicle identities across different camera FOVs. A major challenge in MCVT arises from the small inter-class variability and large intra-class variability among vehicle identities. The same vehicle can appear dramatically different under varying lighting conditions, viewpoints, or occlusions, which complicates cross-camera matching. Conversely, vehicles of the same model often exhibit only subtle appearance differences, making re-identification difficult even for humans \cite{tang2019cityflow}. Luckily, in recent years, many researchers have developed different methods to address these challenges, e.g., \cite{liu2021citywin,qian2020electricity,yang2022citywin,lin2025cityscalemulticameravehicletracking,lin2025ablation}, achieving promising accuracy and establishing a solid theoretical foundation for the MCVT task. However, most existing approaches rely on complex deep neural networks or multi-stage algorithmic pipelines, such as performing computationally heavy offline re-matching after single-camera tracking, with some studies even combining outputs from multiple tracking algorithms to improve accuracy \cite{li2022multi}. Such approaches incur high computational costs and are often limited to offline operations, usually long after a traffic event has occurred. Moreover, when the system needs to scale up to a city-wide deployment with a large number of cameras, the computational burden increases substantially: the intra-camera processing cost increases linearly with the number of cameras, while inter-camera identity association increases quadratically. These challenges limit both the real-time performance and scalability of existing MCVT systems. Overall, there is a substantial gap between academic research and the real-world implementation of MCVT frameworks.

%I'm starting to review here, previous reviews in shared word file - NZ
With the rapid advancement of mobile hardware capabilities and the growing adoption of edge computing technologies, feasible real-time and scalable solutions to the multi-camera vehicle tracking (MCVT) problem are emerging \cite{edgecomputingreview2023,ke2020smart}. Generally, edge computing enables the distribution of computational tasks from a central server to devices located closer to the data source. In this setting, using an edge-based architecture allows the entire intra-camera process to be deployed and executed on a computational unit near the camera \cite{ke2022real}. Specifically, each edge node independently conducts the object detection and single-camera tracking tasks, after which only the processed metadata---such as vehicle IDs, appearance features, bounding box locations, and timestamps---are transmitted to the central server for the subsequent inter-camera association stage \cite{edgecomputingreview2021}. This distributed design offers several advantages: (i) it reduces the central server’s computational burden by distributing most processing across parallel edge nodes, thereby improving scalability; (ii) it lowers communication bandwidth requirements, as transmitting compact metadata is far more efficient than streaming raw video; and (iii) it improves data privacy and storage efficiency by limiting the transmission and storage of sensitive visual content. Together, these benefits provide a strong foundation for real-time, scalable, and privacy-preserving MCVT systems; however, translating this architectural potential into a practical MCVT system still requires task-specific design from both algorithmic and system perspectives. To the best of our knowledge, existing studies have not systematically addressed how to build a real-time and scalable MCVT framework under an edge--server architecture with joint consideration of both aspects. Thus, we consider requirements from multiple perspectives when designing the EASE-MCVT framework:

\textit{(i) Real-time requirements.} Achieving end-to-end real-time performance requires coordinated optimization on both the edge and server sides:
\begin{itemize}
    \item \textit{Edge-side requirements.} To reduce the computational burden on the central server, major intra-camera workloads, particularly object detection and feature extraction, should be shifted to the edge side. At the same time, these modules must remain efficient enough for deployment on resource-constrained edge hardware. 
    
    \item \textit{Server-side requirements.} On the server side, one major challenge is that metadata from different edge nodes may vary in quality or be received asynchronously due to factors such as network latency or frame loss. Therefore, the server-side association process must operate reliably under asynchronous and imperfect input conditions.
\end{itemize}

\textit{(ii) Systems-level scalable deployment requirements.} Enabling real-world city-scale deployment also requires system-level optimization through data-engineering design and deployment standardization:
\begin{itemize}
    \item \textit{Data requirements.} A practical deployment must support structured data exchange between distributed components with low latency, high throughput, and robustness over the network. It also requires stable data interfaces and explicit schemas to ensure reliable communication and reduce integration friction.

    \item \textit{Deployment and maintenance requirements.} City-scale systems require repeatable installation, configuration, updates, and monitoring across heterogeneous hardware and distributed sites. They must also support long-term maintainability so that the system can evolve without disrupting existing components.
\end{itemize}

\textit{(iii) Accuracy requirements.} A reliable city-scale MCVT framework must maintain strong tracking performance under diverse real-world conditions, including viewpoint variation, occlusion, illumination changes, and heterogeneous camera layouts. Therefore, in addition to real-time and deployment considerations, the framework must incorporate task-specific designs that improve the robustness of both intra-camera processing and cross-camera association.

\begin{figure*}[!t]
	\centering 
	\includegraphics[width=0.98\textwidth]{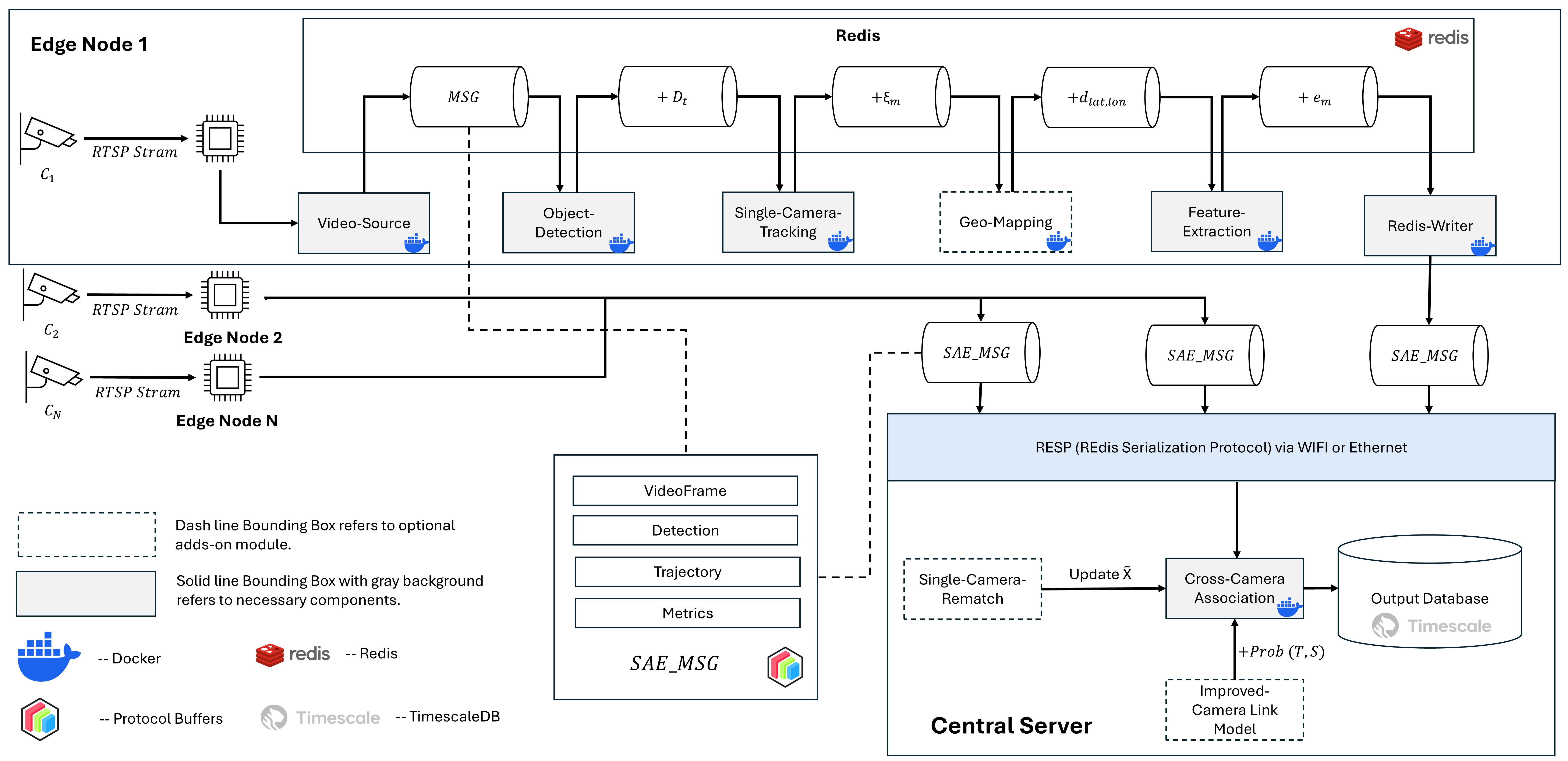}	
    \caption{Overview of the proposed EASE-MCVT edge--server framework. (I) A set of $N$ edge nodes, each connected to one or more cameras, consumes live RTSP streams and performs the intra-camera pipeline (detection, single-camera tracking, geo-mapping, and feature extraction). Only lightweight metadata (tracklets, spatial coordinates, and appearance embeddings) are serialized and transmitted via RESP to (II) the central server, which performs inter-camera association and maintains global multi-camera trajectories.}
	\label{fig:workflow}%
\end{figure*}

Motivated by these requirements, we propose EASE-MCVT, an edge--server MCVT framework designed for real-time operation and city-scale deployment. \cref{fig:workflow} illustrates the overall workflow of EASE-MCVT. The main contributions of this work are as follows:

\begin{itemize}
    \item \textit{Edge--server architecture for deployment-oriented MCVT.} We propose an edge--server architecture for the MCVT task, in which the intra-camera pipeline is executed on edge nodes while the inter-camera association process is handled by a central server. This design enables parallel processing of camera streams, reduces the communication burden through metadata-only transmission, and provides a practical foundation for scalable deployment.

    \item \textit{Algorithmic optimization for real-time and robust operation.} To support real-time throughput and reliable cross-camera tracking, we design a motion-detector trigger for object detection, a dynamic workload scheme for tracklet-level feature extraction, and an online triggering scheme for asynchronous cross-camera association, and incorporate a single-camera re-match module together with a self-supervised camera link model to improve association robustness and accuracy.

    \item \textit{System-level implementation for practical deployment.} To improve deployability, communication efficiency, and long-term maintainability, we integrate production-oriented system components into the framework, including Docker for modular deployment, Redis for low-latency message passing, Protocol Buffers for standardized data exchange, and TimescaleDB for persistent trajectory storage and management.

    \item \textit{Comprehensive framework evaluation.} We conduct extensive experiments to demonstrate the real-time performance and scalability of the proposed framework. We evaluate its cross-camera association accuracy on the \textit{RoundaboutHD} and \textit{CityFlow} datasets and perform ablation studies to validate the effectiveness of the proposed design components.
\end{itemize}

The rest of the paper is structured as follows: Section 2 provides a comprehensive literature review of the multi-camera tracking problem and researchers' efforts to develop a real-time multi-camera tracking pipeline. Section 3 outlines the algorithmic details of the EASE-MCVT framework, while Section 4 presents evaluations on different datasets. Finally, Section 5 summarises the main findings and proposes directions for future research.
\section{Background}
As already noted, the MCVT problem can be divided into two stages: (i) the intra-camera process and (ii) the inter-camera process. The intra-camera process aims to detect vehicles and maintain consistent identities across frames from a single camera \cite{amosa2023review}. \textit{(1) Vehicle Detection.} Current deep-learning-based methods can be mainly classified into two categories: one-stage detectors, such as the YOLO family \cite{tian2025yolov12,varghese2024yolov8}, and two-stage detectors, such as Fast R-CNN \cite{girshick2015fastrcnn}. Since 2020, one-stage detectors---particularly the YOLO family---have become dominant in both academic research and real-world applications due to their balance of speed and accuracy. More recently, some researchers have also attempted to adopt powerful pre-trained multimodal large language models (MLLMs) for zero-shot object detection \cite{zang2025vlm4objdetection,bai2023qwen}. Due to the large size and corresponding computational cost of MLLMs, the YOLO family remains the first choice for edge-based real-world applications. \textit{(2) Single-Camera Tracking.} After acquiring object locations, the single-camera tracking module aims to maintain the temporal consistency of the same objects. Existing SCT methods can be broadly categorized into two groups: deep-embedding-based and non-embedding-based methods. For deep-embedding-based algorithms, e.g., DeepSORT \cite{deepsort} and BotSort \cite{bot-sort}, the similarity of objects across different frames depends on both bounding-box (Bbox) Intersection over Union (IoU) and deep-embedding similarity. The use of deep embeddings improves the robustness of the tracker in handling occlusion and frame loss but introduces additional computational cost into the SCT process. Non-embedding-based methods, such as SORT \cite{sort} and ByteTrack \cite{zhang2022bytetrack}, rely only on positional prediction and matching, which is lightweight and more suitable for resource-constrained, edge-based real-world applications. \textit{(3) Feature Extraction.} In an edge--server MCVT setting, appearance features are typically computed on the edge and sent to the server to support cross-camera association. Once detections are produced and temporally linked by SCT, the next step is to compute compact and discriminative appearance features that can be sent to the downstream cross-camera association module. One of the biggest challenges for feature extraction is the need for substantial computational resources to obtain robust deep embeddings. Many existing approaches \cite{yang2022citywin,li2022multi} rely on complex CNN backbones such as ResNet \cite{he2016deepresnet} to extract discriminative features. Some offline pipelines further increase complexity by combining multiple backbones (e.g., Transformers and several ResNet variants) to obtain stronger embeddings \cite{zhang2024video}, which is impractical for real-time deployment on embedded hardware.

The inter-camera process refers to linking vehicle identities across different cameras. Depending on how an identity is represented, this linking can be formulated at two levels. In image-based vehicle ReID, each identity is represented by one or more still images captured by different cameras, and the goal is to match these image instances across views. In video-based ReID, which is MCVT, each identity is represented by a sequence of observations produced by the intra-camera pipeline (e.g., tracklets or trajectories), and the goal is to associate these trajectories across cameras to form continuous cross-camera vehicle tracks. Image-based ReID typically ignores temporal continuity, which simplifies the matching problem but decouples performance from intra-camera tracking quality. Consequently, image-based ReID focuses on extracting representative deep embeddings and applying different matching methods. Some well-known studies include PROVID \cite{liu2017provid}, Git \cite{shen2023git}, and Going Beyond Real Data \cite{zheng2020going}. These studies provide a solid foundation for the inter-camera MCVT process. As computer vision technologies have become more mature, researchers have increasingly focused on the video-based vehicle ReID problem. Compared with image-based ReID, video-based settings require processing many more observations per identity, which provides more visual information but also increases the complexity of the problem. To mitigate these challenges, some researchers \cite{yang2022citywin} introduce heavy SCT modules. For example, \cite{li2022multi} combines the results from three different SCT modules, which introduces a substantial computational load. In addition, the SCT module in \cite{yang2022citywin} takes computationally heavy deep embeddings as input, combines multiple SCT algorithms, and introduces an offline re-linking module, making the SCT module over-engineered. Other studies leverage manually defined zones to encode spatio-temporal relationships within and across adjacent cameras \cite{liu2021citywin,wu2021multi}. While such zone-based constraints have proven effective in controlled or small-scale environments, they are difficult to scale: zone definitions require manual effort, and any change in camera placement or orientation may necessitate re-labeling. Overall, existing work highlights a central challenge for practical deployment---achieving a reliable balance between tracking accuracy, computational efficiency, and system scalability.

Although no existing system fully addresses the real-world Multi-Camera Vehicle Tracking (MCVT) problem, several studies have made progress toward enabling real-time and deployable vision-based ITS. For instance, Zhang and Izquierdo \cite{zhang2019real} propose a real-time multi-camera tracking framework that processes video data frame by frame instead of waiting for each batch to complete, achieving real-time performance on the DukeMTMC dataset \cite{dukemtmct}. More recently, Zhuang \textit{et al.} \cite{zhuang2024real} developed a real-time MCVT framework that takes continuous video streams as input and demonstrates real-time performance on both the CityFlow dataset \cite{tang2019cityflow} and their custom dataset. However, the real-time capability of these systems depends heavily on high-performance hardware, and computational performance quickly becomes a bottleneck as the number of cameras increases. To mitigate this limitation, several studies \cite{liu2023realedge, ke2020smart} have explored edge--server architectures that enable the distribution of computational workloads for practical ITS deployment. Although these systems are not designed specifically for MCVT, their real-world deployments provide evidence that edge-based architectures can improve scalability for vision-driven ITS applications. Another closely related research direction is cooperative edge-based vehicle ReID. For instance, \cite{yang2023cooperative} developed an edge-enabled vehicle ReID system deployed under real traffic conditions; however, it operates at the clip level rather than maintaining full trajectories. The work most closely related to EASE-MCVT is that of Gaikwad and Karmakar \cite{gaikwad2021smart}, which applies edge computing to multi-camera pedestrian tracking to achieve scalable and real-time performance. However, in real traffic networks, pedestrians exhibit much lower speeds, smaller activity ranges, and smaller scales than vehicles, making vehicle tracking a far more challenging problem.
\section{EASE-MCVT Framework}
\subsection{EASE-MCVT Overview}

\textbf{Problem Definition:} Consider a camera network with $N$ cameras indexed by $\mathcal{C}=\{1,\dots,N\}$. Each camera $i\in\mathcal{C}$ provides a video stream (or a set of clips) over a time interval $[t_1,t_2]$. Let $\mathcal{T}$ denote the discrete timestamp set within $[t_1,t_2]$ (e.g., frame times). At each time $t\in\mathcal{T}$, the detector produces a set of vehicle detections $\mathcal{D}_{i,t}=\{d_{i,t}^{(k)}\}_{k=1}^{n_{i,t}}$, where each detection is represented by metadata $d_{i,t}^{(k)}=\bigl(t,\mathbf{b}_{i,t}^{(k)},s_{i,t}^{(k)}\bigr)$. Here $\mathbf{b}_{i,t}^{(k)}\in\mathbb{R}^4$ is the bounding box (e.g., $(x,y,w,h)$) and $s_{i,t}^{(k)}\in[0,1]$ is the detection confidence. After we get the detection, the single camera tracker will group detections into a set of local tracklets (trajectories) for each camera, denoted by $\Xi_i=\{\xi_{i,m}\}_{m=1}^{M_i}$ for $i\in\mathcal{C}$. Each local tracklet $\xi_{i,m}$ is an ordered sequence of detections $\xi_{i,m}=\{ d_{i,t}^{(k)} \mid t\in\mathcal{T}_{i,m}\subseteq\mathcal{T} \}$ and is associated with a \emph{local} identity label $\ell_{i,m}$ that is only meaningful within camera $i$. Notably, in our framework, each local tracklet is further represented by a trajectory-level appearance embedding $\mathbf{e}_{i,m}\in\mathbb{R}^d$, yielding the augmented tracklet representation $\tilde{\xi}_{i,m}=\bigl(\xi_{i,m},\mathbf{e}_{i,m}\bigr)$. Collecting all augmented tracklets across the camera network, we define $\tilde{\Xi}_i=\{\tilde{\xi}_{i,m}\}_{m=1}^{M_i}$ and $\tilde{\mathcal{X}}=\bigcup_{i\in\mathcal{C}}\tilde{\Xi}_i$ as the set of all augmented local tracklets observed during $[t_1,t_2]$. The goal of MCVT is then to associate these augmented tracklets across cameras and recover \emph{global} vehicle identities by estimating a global identity assignment $\phi:\tilde{\mathcal{X}}\rightarrow\mathcal{G}$, where $\mathcal{G}$ is the set of global vehicle identities, such that for any two augmented tracklets $\tilde{\xi}_{i,m},\tilde{\xi}_{j,n}\in\tilde{\mathcal{X}}$, $\phi(\tilde{\xi}_{i,m})=\phi(\tilde{\xi}_{j,n})$, if and only if $\tilde{\xi}_{i,m}$ and $\tilde{\xi}_{j,n}$ correspond to the same physical vehicle. The resulting global trajectory set can be written as $\mathcal{T}^{\text{global}}=\{\mathcal{T}_g\}_{g\in\mathcal{G}}$ with $\mathcal{T}_g=\{\tilde{\xi}\in\tilde{\mathcal{X}}\mid \phi(\tilde{\xi})=g\}$, i.e., each global trajectory $\mathcal{T}_g$ is a cross-camera collection of local tracklets which will share the same global identity.

\textbf{Framework Architecture.} We design EASE-MCVT to solve the MCVT problem defined above while meeting the practical requirements of real-time throughput and city-scale scalability. From experience, most of the computational load happens in the intra-camera process; as the number of cameras increases, executing all intra-camera processing on a single central machine quickly becomes infeasible. This motivates an edge–server architecture in which each camera stream is processed locally on an edge device (NVIDIA Jetson AGX Orin in our deployment), and only lightweight metadata is transmitted to a central server for inter-camera association. An overview of the architecture is shown in \cref{fig:workflow}. On the edge side, each camera connects directly to its corresponding edge node, which consumes the RTSP stream and converts it into a structured format via the \textit{video-source} module. The stream is then processed by \textit{object detection} and \textit{single camera tracking} to generate local tracklets, followed by \textit{geo-mapping} and \textit{feature extraction} to attach geo-coordinates and trajectory-level appearance embeddings to each tracklet. The resulting messages are serialized and published by the \textit{Redis-Writer} to the central server over network. On the server side, we perform \textit{cross-camera association} to link tracklets from different cameras and assign global identities. To further improve robustness and accuracy, we also include two optional modules: \textit{single-camera rematch} and an \textit{improved camera-link model}. Finally, the global identities $\mathcal{G}$ and associated trajectories are stored in a database. The details of each module are described in the rest of this section.

% Try to re-organize the story telling structure in further revision. I want to explain the pipeline in one consistent narrative.

\subsection{Edge Node Algorithm Design}
\subsubsection{Video-Source Raw Video Pre-Process}
To facilitate downstream intra-camera processing, we design a \textit{video-source} module to pre-process the incoming RTSP stream. For each frame, the module (i) captures the image, (ii) attaches basic metadata including the camera/source ID, timestamp, and frame resolution, and (iii) encodes the frame payload as JPEG-compressed bytes for efficient transport. The resulting frame message is then structured using a protobuf and serialized for transmission to subsequent modules in the edge pipeline.
\subsubsection{Real-time Object Detection}
\paragraph{YOLO detector}
In recent years, the YOLO family of detectors has become dominant in both academic research and real-world applications due to their speed and accuracy. For our deployment, we evaluated three recent YOLO variants pre-trained on the COCO dataset \cite{lin2014microsoft}: YOLOv8 \cite{varghese2024yolov8}, YOLOv11 \cite{khanam2024yolov11}, and YOLOv12 \cite{tian2025yolov12}, which differ in architecture and computational cost. Through our inspection, all three models exhibited frequent missed detections for vehicles viewed from an overhead perspective, as well as for small and distant vehicles. To address this domain gap, we fine-tuned these YOLO models on the RoundaboutHD object-detection dataset \cite{lin2025roundabouthd}. After comparing detection accuracy and inference latency, we selected the fine-tuned YOLOv11n model and converted it to TensorRT \cite{NVIDIA_TensorRT_Docs} to enable more efficient inference on the edge device.
\paragraph{Motion Detector}
We observe that traffic scenes often include extended intervals with little or no vehicle activity—particularly during late-night hours—making it inefficient to run the full detection and tracking pipeline continuously on edge hardware. To reduce unnecessary computation and power consumption, we introduce a lightweight motion detector placed upstream of the YOLO module to filter those near empty scenes. For efficiency, we adopt the classical MOG2 background-subtraction method \cite{mog2} and estimate the number of foreground (moving) pixels per frame. The activation threshold $\tau_{\mathrm{fg}}$ is set as a fixed fraction of the image area, which normalizes the criterion across different resolutions. If the foreground pixel count exceeds $\tau_{\mathrm{fg}}$, we activate the YOLO-based detector; otherwise, the edge pipeline skips downstream processing for that frame.

\subsubsection{Single Camera Tracking}
In the SCT module, detections from each incoming frame are associated with existing in-memory tracklets to maintain temporal identity consistency. Using the open-source, pluggable tracking library BOXMOT \cite{brostrom2025boxmot}, we evaluated several state-of-the-art trackers, including both embedding-based and non-embedding-based methods, in terms of FPS and IDF1. Based on these results and the trade-off between runtime efficiency and robustness on edge hardware, we select ByteTrack \cite{zhang2022bytetrack} as the default SCT algorithm in EASE-MCVT. ByteTrack is embedding-free and therefore avoids per-object feature extraction overhead, which is beneficial on resource-constrained edge devices. We also keep the interface modular so that alternative trackers can be swapped in when different operating conditions or accuracy–latency requirements apply.
\subsubsection{Geo-Mapping}
The geo-mapping module converts the pixel-space location of each detection into a real-world position, expressed either in GPS coordinates (latitude, longitude, elevation) or in a camera space coordinate system. This spatial information provides additional evidence for multi-camera association, especially when cameras have overlapping or partially overlapping fields of view (FOV). The mapping is defined by a camera transformation model parameterized by camera specifications (e.g., field of view, heading, tilt, roll, and mounting height) and the camera’s geo-location. In practice, these parameters can be obtained from sensor specifications and site measurements, or calibrated using map priors (e.g., from satellite imagery) when the camera location is known. We implement this transformation using the camera model provided by Gerum \textit{et al.} \cite{gerum2017cameratransform}:
\begin{equation}
\label{eq:camera_transform}
\mathbf{p}^{w}_{i,t} =
\mathrm{Cam}\!\left(\,x^{\mathrm{pix}}_{i,t},\, y^{\mathrm{pix}}_{i,t},\, h_{i,t};\, \boldsymbol{\theta}_i \right),
\end{equation}
where $\mathbf{p}^{w}_{i,t} = [\,\mathrm{lat}_{i,t},\, \mathrm{lon}_{i,t},\, z_{i,t}\,]^\top$ denotes the estimated world position of the vehicle at time $t$ in camera $i$, $(x^{\mathrm{pix}}_{i,t}, y^{\mathrm{pix}}_{i,t})$ is the detection center in pixel coordinates, $h_{i,t}$ is a height prior used to resolve depth, $\boldsymbol{\theta}_i$ denotes the camera parameters, and $\mathrm{Cam}(\cdot)$ is the camera transformation model.
\subsubsection{Feature Extraction}
\paragraph{Tracklet-Level Feature Representation}
Running deep-neural-network based feature extraction on every frame can be computationally demanding on resource-constrained edge devices. To reduce this overhead while maintaining robust appearance feature for downstream association, EASE-MCVT performs feature extraction at the tracklet level rather than the frame level. As a result, the edge node buffers the per-frame tracklet updates and triggers feature extraction after a tracklet is completed within the camera’s field of view.

We adopt a lightweight ResNeXt-50 backbone \cite{resnext} and fine-tuned it on RoundaboutHD ReID dataset to keep the balance between accuracy and inference speed on edge hardware. For a completed local tracklet $\xi_{i,m}$, the feature extractor outputs a matrix of frame-level embeddings $\{\mathbf{e}_{i,m}^{(k)}\}_{k=1}^{N_{i,m}}$, where $\mathbf{e}_{i,m}^{(k)}\in\mathbb{R}^{2048}$ and $N_{i,m}$ is the number of frames selected for feature extraction from this tracklet. We aggregate these embeddings using confidence-weighted averaging followed by $\ell_2$ normalization to obtain a tracklet-level embedding:
\begin{equation}
\label{eq:l2_norm}
\hat{\mathbf{e}}_{i,m}=
\frac{\sum_{k=1}^{N_{i,m}} c_{i,m}^{(k)}\,\mathbf{e}_{i,m}^{(k)}}
{\left\|\sum_{k=1}^{N_{i,m}} c_{i,m}^{(k)}\,\mathbf{e}_{i,m}^{(k)}\right\|_2},
\end{equation}
where $\hat{\mathbf{e}}_{i,m}\in\mathbb{R}^{2048}$ is the normalized tracklet embedding and $c_{i,m}^{(k)}$ is the detection confidence of the $k$-th cropped instance used as the aggregation weight. After aggregation, the edge node only transmits lightweight metadata, which include detection results, frame level embeddings $\{\mathbf{e}_{i,m}^{(k)}\}_{k=1}^{N_{i,m}}$, and tracklet-level embeddings $\hat{\mathbf{e}}_{i,m}$---to the central server for inter-camera association, without sending raw images.
\paragraph{Dynamic Workload Scheme}
Although tracklet-level extraction reduces computation compared with frame-by-frame extraction, the feature-extraction workload still varies with traffic density and tracklet length. In particular, long tracklets can introduce a large number of cropped instances and temporarily increase the processing load. To limit the workload contributed by each completed tracklet, we introduce an adaptive temporal subsampling factor $K_{i,m}$.

For camera \(i\), we define the total number of cropped instances to be processed within a time window \(\Delta\) as
\begin{equation}
\label{eq:datavolume}
V_i(\Delta)
=
\sum_{m=1}^{n_i(\Delta)}
\left\lceil
\frac{\lvert\mathcal{T}_{i,m}\rvert}{K_{i,m}}
\right\rceil,
\end{equation}
where \(n_i(\Delta)\) is the number of tracklets completed on camera \(i\) during \(\Delta\), \(\mathcal{T}_{i,m}\) denotes the set of frames in tracklet \(\xi_{i,m}\), and \(\lvert\mathcal{T}_{i,m}\rvert\) is its length in frames. A larger \(K_{i,m}\) results in stronger temporal subsampling and therefore reduces the number of cropped instances processed for the corresponding tracklet.

To maintain real-time throughput while preserving as many visual observations as possible, we use the smallest subsampling factor that satisfies a predefined per-tracklet processing budget. Specifically, we define
\begin{equation}
\label{eq:dws}
K_{i,m}
= \max\!\left(1,\;
\left\lceil
\frac{\lvert\mathcal{T}_{i,m}\rvert}{Q_{\max}(\Delta)}
\right\rceil
\right),
\end{equation}
where \(Q_{\max}(\Delta)\) is a hardware-dependent hyperparameter specifying the maximum number of cropped instances selected from an individual tracklet for processing within the target time window \(\Delta\). It is empirically calibrated for the target edge hardware prior to deployment to provide sufficient processing margin and avoid excessive queue accumulation under the evaluated workload.

Under this policy, shorter tracklets retain dense visual observations, whereas longer tracklets are subsampled more aggressively. Consequently, the processing workload contributed by an individual tracklet is bounded by the predefined budget \(Q_{\max}(\Delta)\), preventing unusually long tracklets from dominating the feature-extraction queue. Since multiple tracklets may complete within the same time window, this per-tracklet budget does not impose a strict upper bound on the aggregate workload \(V_i(\Delta)\). Instead, system-level runtime stability under varying traffic loads is evaluated experimentally using queue occupancy and overflow statistics.
\subsection{Central Server Design}
\subsubsection{Single-Camera Re-Match}
SCT in the intra-camera stage can be degraded by illumination changes, occlusions, irregular vehicle motion, and occasional frame loss. These effects often lead to track identity switches, especially when lightweight, non-embedding-based trackers are used on edge devices. To reduce the impact of imperfect SCT on downstream cross-camera association without increasing edge-side computation, we introduce a \textit{single-camera re-match} module on the central server, where greater compute resources are available. This module refines the SCT output by re-linking fragmented tracklets that are likely to belong to the same vehicle within the same camera. For the re-match process, we apply a rule-based merging policy. Given two candidate tracklets $\text{Tracklet}_A$ and $\text{Tracklet}_B$ from the same camera, where $\text{Tracklet}_A$ temporally precedes $\text{Tracklet}_B$, we merge them if and only if the following three conditions are all satisfied:
\begin{equation}
\label{eq:merge_condition}
\begin{aligned}
&\mathrm{Merge}(\text{Tracklet}_A,\text{Tracklet}_B)\\
&= \mathbb{I}\Big[
\Delta t < T_{\mathrm{th}}
\ \wedge\
\Delta c < D_{\mathrm{th}}
\ \wedge\
\Delta e < F_{\mathrm{th}}
\Big].
\end{aligned}
\end{equation}
where $\mathbb{I}[\cdot]$ is an indicator function, and
\begin{equation}
\label{eq:merge_terms}
\begin{split}
\Delta t &= t_B^{\mathrm{start}} - t_A^{\mathrm{end}}, \\
\Delta c &= \left\|\mathbf{c}_B^{\mathrm{start}} - \mathbf{c}_A^{\mathrm{end}}\right\|_2, \\
\Delta e &= 1 - \max\Big(
    \operatorname{cos\_sim}(\bar{\mathbf{e}}_A,\bar{\mathbf{e}}_B),\,
    \operatorname{cos\_sim}(\mathbf{e}_A^{\mathrm{end}},\mathbf{e}_B^{\mathrm{start}})
\Big).
\end{split}
\end{equation}

Here $t_A^{\mathrm{end}}$ and $t_B^{\mathrm{start}}$ denote the end time of $\text{Tracklet}_A$ and the start time of $\text{Tracklet}_B$, respectively, and $T_{\mathrm{th}}$ is the maximum allowed temporal gap. The vectors $\mathbf{c}_A^{\mathrm{end}},\,\mathbf{c}_B^{\mathrm{start}}\in\mathbb{R}^2$ are the bounding-box centers at the end of $\text{Tracklet}_A$ and the start of $\text{Tracklet}_B$, and $D_{\mathrm{th}}$ is the spatial distance threshold. The vectors $\bar{\mathbf{e}}_A,\,\bar{\mathbf{e}}_B\in\mathbb{R}^d$ are the $\ell_2$-normalized averages of the per-detection embeddings within each tracklet, while $\mathbf{e}_A^{\mathrm{end}}$ and $\mathbf{e}_B^{\mathrm{start}}$ are the embeddings of the last detection of $\text{Tracklet}_A$ and the first detection of $\text{Tracklet}_B$. Finally, $\operatorname{cos\_sim}(\cdot,\cdot)$ computes cosine similarity and $F_{\mathrm{th}}$ is the appearance-distance threshold.

We adopt a conservative merging strategy by using tight thresholds ($T_{\mathrm{th}}, D_{\mathrm{th}}, F_{\mathrm{th}}$), prioritizing precision over recall. This choice is because we believe an incorrect re-merge irreversibly corrupts the identity history and can mislead downstream cross-camera association, whereas leaving tracklets unmerged is less harmful. The updated tracklets set $\tilde{\mathcal{X}}$ are then forwarded to the \textit{cross-camera association} module.

\subsubsection{Self-Supervised Camera Link Model}
To improve cross-camera association accuracy, many prior studies (e.g., \cite{hsu2019multi}) introduce the spatial-temporal constraints derived from road topology and travel-time statistics to filter infeasible matches and narrow down the searching space when linking tracklets across adjacent cameras. While effective, manually defining such constraints is labor-intensive and becomes impractical at city scale. Following on from our previous work \cite{lin2025cityscalemulticameravehicletracking}, we adopt a \textit{self-supervised camera link model (CLM)} that learns camera-to-camera spatial-temporal relationships offline without requiring human annotations. The model consists of three major steps.

\textbf{(1) Identifying entry and exit zones within each camera.} 
We first estimate a set of entry and exit zones for each camera view. Intuitively, entry zones correspond to regions where vehicles frequently appear (i.e., where tracklets start), while exit zones correspond to regions where vehicles frequently disappear (i.e., where tracklets end). For camera $i$, let $\mathcal{Z}_i$ denote the set of all discovered zones, with entry zones $\mathcal{Z}_i^{\mathrm{entry}}\subseteq \mathcal{Z}_i$ and exit zones $\mathcal{Z}_i^{\mathrm{exit}}\subseteq \mathcal{Z}_i$. Zones are obtained in two stages: (i) we cluster the start pixel locations and end pixel locations of all tracklets in camera $i$; (ii) for each cluster, we define the zone region by fitting an axis-aligned bounding box that encloses the cluster points. Each fitted region is treated as a candidate zone. Finally, we label each zone as an entry or exit zone according to the entry/exit density of tracklet starts and ends within that region.

\textbf{(2) Pairing entry--exit zones across adjacent cameras.} 
After obtaining entry and exit zones for each camera, the next step is to establish correspondences between zones in adjacent camera pairs. For an adjacent camera pair $(i,j)$, let $\mathcal{Z}_i^{\mathrm{exit}}=\{\mathbf{z}_{i,m}^{\mathrm{exit}}\}_{m=1}^{N_i^{\mathrm{exit}}}$ and $\mathcal{Z}_j^{\mathrm{entry}}=\{\mathbf{z}_{j,n}^{\mathrm{entry}}\}_{n=1}^{N_j^{\mathrm{entry}}}$. We compute a zone-pairing score for each candidate pair using the $\mathrm{ZonePair}(\cdot)$ function in \cite{lin2025cityscalemulticameravehicletracking}:
\begin{equation}
s_{i,j}\!\left(\mathbf{z}_{i,m}^{\mathrm{exit}},\,\mathbf{z}_{j,n}^{\mathrm{entry}}\right)
= \mathrm{ZonePair}\!\left(\mathbf{z}_{i,m}^{\mathrm{exit}},\,\mathbf{z}_{j,n}^{\mathrm{entry}}\right),
\quad \forall\, m,n.
\end{equation}
We then select the best-matching correspondence by
\begin{equation}
(\hat m,\hat n)
= \arg\max_{m ,\, n}
s_{i,j}\!\left(\mathbf{z}_{i,m}^{\mathrm{exit}},\,\mathbf{z}_{j,n}^{\mathrm{entry}}\right).
\end{equation}

We apply the same procedure in the reverse direction (i.e., pairing $\mathcal{Z}_j^{\mathrm{exit}}$ with $\mathcal{Z}_i^{\mathrm{entry}}$). As a result, each adjacent camera pair yields directional entry--exit correspondences that serve as spatial constraints for downstream association: when matching tracklets from camera $i$ to camera $j$, we only consider candidates that end in the selected exit zone of camera $i$ and begin in the selected entry zone of camera $j$.

\textbf{(3) Camera-link transition-time estimation.} 
For each directional paired zone correspondence $(i\!\rightarrow\! j)$ (i.e., an exit zone in camera $i$ paired with an entry zone in camera $j$), we estimate the distribution of vehicle transition times using Gaussian kernel density estimation (KDE). 

\emph{Transition-time sample collection.} Following Lin \textit{et al.} \cite{lin2025cityscalemulticameravehicletracking}, we first construct a set of high-confidence cross-camera matches offline to obtain reliable transition-time samples. Specifically, we select candidate tracklets that (i) end in the paired exit zone of camera $i$ and (ii) begin in the paired entry zone of camera $j$. We then perform a preliminary cross-camera association using strict appearance thresholds to favor precision (i.e., minimizing false matches). The resulting matched tracklet pairs are treated as pseudo ground truth, and for each matched pair we compute the transition time as
\begin{equation}
\label{eq:tau_def}
\tau_{i,j}^{(\ell)} = t_{j,n}^{\mathrm{start}} - t_{i,m}^{\mathrm{end}},
\end{equation}
where $t_{i,m}^{\mathrm{end}}$ is the end time of the matched tracklet in camera $i$ and $t_{j,n}^{\mathrm{start}}$ is the start time of the matched tracklet in camera $j$. Collecting all samples yields $\{\tau_{i,j}^{(\ell)}\}_{\ell=1}^{L_{i,j}}$.

\emph{KDE estimation.} Given the observed samples, we estimate a continuous transition-time density:
\begin{equation}
\label{eq:kde}
\hat p_{i,j}(t)
= \frac{1}{L_{i,j} h}
  \sum_{\ell=1}^{L_{i,j}}
  \mathcal{K}\!\left(
    \frac{t - \tau_{i,j}^{(\ell)}}{h}
  \right),
\end{equation}
where $h>0$ is the bandwidth and $\mathcal{K}(\cdot)$ is the Gaussian kernel
\begin{equation}
\mathcal{K}(u)
= \frac{1}{\sqrt{2\pi}}
  \exp\!\left(-\frac{u^{2}}{2}\right).
\end{equation}

The learned density $\hat p_{i,j}(t)$ provides a temporal constraint for online association. In summary, the self-supervised camera link model allows us to restrict candidate cross-camera association to those that (i) satisfy the learned zone correspondence and (ii) have transition times that are sufficiently likely under $\hat p_{i,j}(t)$.

\subsubsection{Cross-Camera Association}
Cross-camera association links tracklets from different cameras that correspond to the same physical vehicle. We perform association over the road-topology graph: for each topologically adjacent camera pair $(i,j)$, we construct a pairwise cost matrix and solve a one-to-one assignment. Applying the same procedure to all adjacent pairs yields global identities across the network.

\textbf{Cost matrix construction.} For an adjacent camera pair $(i,j)$, let $\tilde{\Xi}_i=\{\tilde{\xi}_{i,m}\}_{m=1}^{M_i}$ and $\tilde{\Xi}_j=\{\tilde{\xi}_{j,n}\}_{n=1}^{M_j}$ denote the augmented tracklet sets from cameras $i$ and $j$, respectively, where each $\tilde{\xi}_{i,m}=(\xi_{i,m},\mathbf{e}_{i,m})$ consists of a local tracklet and its trajectory-level embedding. For a candidate pair $(\tilde{\xi}_{i,m},\tilde{\xi}_{j,n})$, we define the travel-time gap as
\begin{equation}
\label{eq:delta_t}
\Delta t_{m,n} = t^{\mathrm{start}}(\xi_{j,n}) - t^{\mathrm{end}}(\xi_{i,m}),
\end{equation}
and compute the pairwise association cost as

\begin{equation}
\label{eq:association_cost}
\mathrm{Cost}
\left(
\tilde{\xi}_{i,m},
\tilde{\xi}_{j,n}
\right)
=
\delta
\left[
1-\mathrm{sim}
\left(
\mathbf{e}_{i,m},
\mathbf{e}_{j,n}
\right)
\right]
-
\epsilon
\hat{p}_{i,j}
\left(
\Delta t_{m,n}
\right).
\end{equation}

where $\operatorname{sim}(\cdot,\cdot)$ denotes cosine similarity between embeddings, $\hat{p}_{i,j}(\cdot)$ is the transition-time density learned by the camera-link model for the directional link $(i\!\rightarrow\! j)$, and $\delta$ and $\epsilon$ weight the appearance and temporal terms, respectively. Candidate pairs that violate the learned zone-based spatial constraint will be filtered before cost construction when it is available.

The full cost matrix between cameras $i$ and $j$ is
\begin{equation}
\label{eq:cost_matrix}
\mathbf{C}_{i,j} =
\begin{bmatrix}
\mathrm{Cost}(\tilde{\xi}_{i,1}, \tilde{\xi}_{j,1}) & \cdots & \mathrm{Cost}(\tilde{\xi}_{i,1}, \tilde{\xi}_{j,M_j}) \\
\vdots & \ddots & \vdots \\
\mathrm{Cost}(\tilde{\xi}_{i,M_i}, \tilde{\xi}_{j,1}) & \cdots & \mathrm{Cost}(\tilde{\xi}_{i,M_i}, \tilde{\xi}_{j,M_j})
\end{bmatrix}.
\end{equation}

\textbf{Cost matrix solving and online triggering.} For each adjacent camera pair $(i,j)$, we solve a one-to-one assignment on $\mathbf{C}_{i,j}$ to associate tracklets that likely correspond to the same physical vehicle. We adopt greedy matching as default due to its low latency: candidate pairs are ranked by $\mathrm{Cost}(\cdot,\cdot)$ in ascending order, and matches are accepted sequentially while enforcing the one-to-one constraint. We also provide the Hungarian algorithm as an alternative when an optimal assignment is preferred. To avoid spurious associations, we reject matched pairs whose cost is higher than a predefined threshold. After solving all adjacent camera pairs, we consolidate the resulting matches to update the global identity assignment $\phi:\tilde{\mathcal{X}}\rightarrow\mathcal{G}$ and the corresponding global identity set $\mathcal{G}$.

In principle, cross-camera association could be triggered whenever the tracklet set $\tilde{\mathcal{X}}$ is updated. However, solving assignments for every incremental update introduces unnecessary overhead and is not required in real-world deployments. Instead, we run association periodically every $\alpha$ frames to balance computational efficiency and end-to-end latency. The results of each association are written to the database, where they are merged with historical records to extend existing trajectories and register newly formed trajectories.

% NOTE: I SHOULD ADD THE EASE_ENGINE PART IN HERE.
\subsection{Data Engineering Implementation}
To support real-world deployment rather than simulation-only evaluation, we implement EASE-MCVT through adopting our industrial partner’s SAE-ENGINE pipeline \cite{starwit_awareness_engine}, which provides production-oriented frameworks including \textit{Docker}, \textit{Redis}, \textit{Protocol Buffers (Protobuf)}, and \textit{TimescaleDB}. These components standardize deployment, data transmission, and storage across heterogeneous edge and server hardware.

Specifically, we use \textit{Docker} to containerize each module as a docker image, enabling reproducible builds, isolated dependencies, and portable deployment across different devices. For inter-module communication, we use \textit{Redis} as a lightweight, low-latency in-memory data store and message broker to support fast reads/writes and efficient message passing between modules. We use \textit{Protobuf} to define a strict schema for all exchanged messages and to serialize metadata compactly, improving interoperability and reducing transmission overhead. Finally, we store association outputs and trajectory records in \textit{TimescaleDB}, a time-series database built on PostgreSQL, to support persistent storage, efficient time-based queries, and long-term data management.
\section{Experiment}
\subsection{Dataset}
\subsubsection{RoundaboutHD}
RoundaboutHD, introduced in our previous work \cite{lin2025roundabouthd}, is a real-world multi-camera vehicle tracking dataset designed for roundabout traffic scenarios. It contains 40 minutes of fully annotated, non-overlapping multi-camera video captured from four 4K cameras at 15\,FPS, with the geographic layout shown in~\cref{fig:roundaboutHD}. The dataset introduces several challenges for MCVT, including (1) nonlinear vehicle trajectories, (2) frequent occlusions caused by infrastructure and other vehicles, and (3) complex environments such as multiple exits and intersections across camera views. At the time of submission, we are not aware of any published work that has fully addressed these challenges. In this work, we use RoundaboutHD as the primary dataset to evaluate our proposed EASE-MCVT framework. 

\subsubsection{CityFlow}
\begin{figure*}[!t]
    \centering
    % Left image
    \begin{subfigure}[b]{0.30\linewidth}
        \centering
        \includegraphics[width=\linewidth]{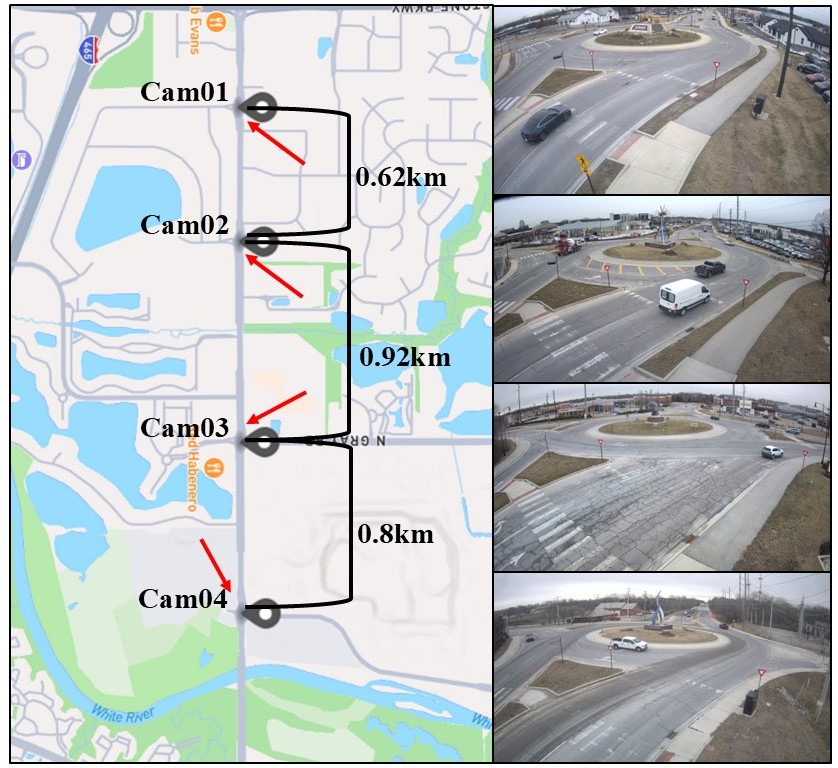} % or .png/.jpg
        \caption{RoundaboutHD dataset.}
        \label{fig:roundaboutHD}
    \end{subfigure}
    \hspace{0.01\linewidth} % try 0.00–0.03
    % Right image
    \begin{subfigure}[b]{0.50\linewidth}
        \centering
        \includegraphics[width=\linewidth]{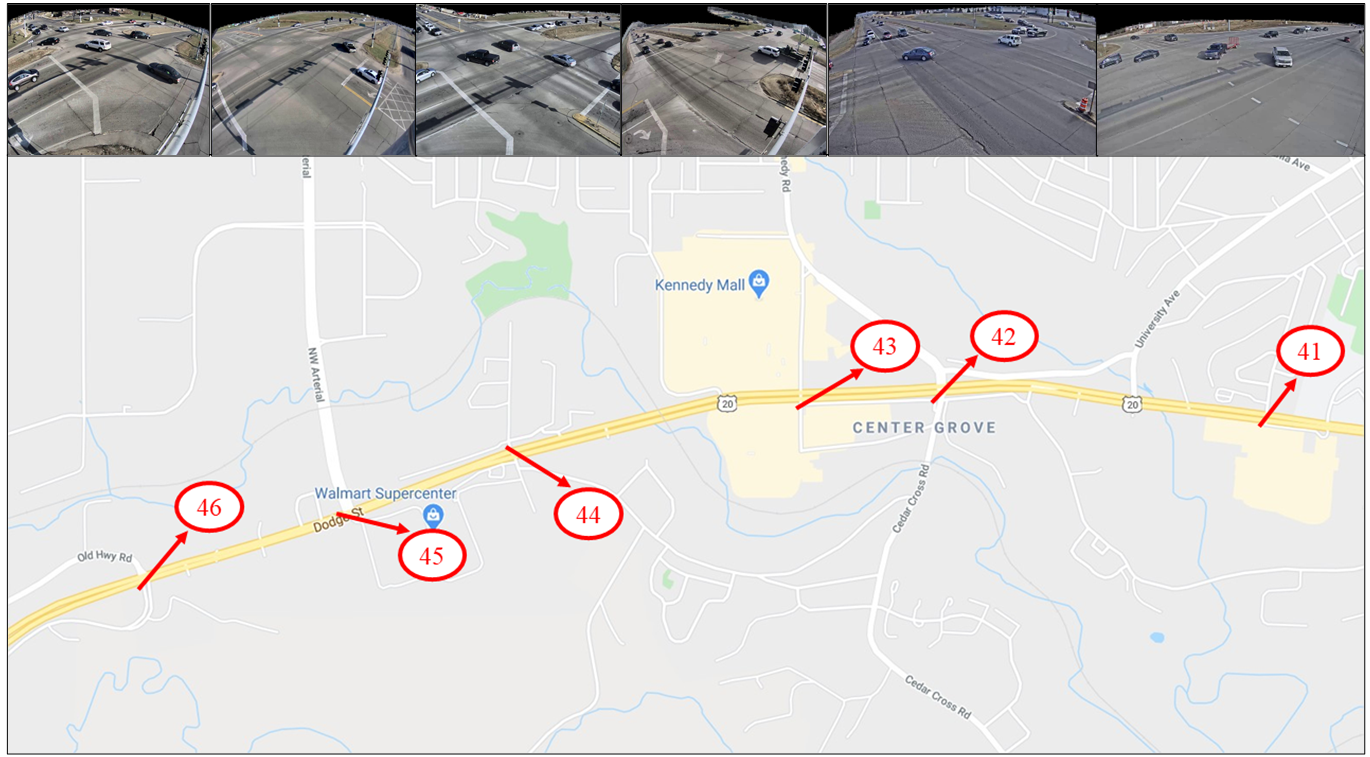}
        \caption{CityFlow test dataset.}
        \label{fig:cityflow}
    \end{subfigure}
    \caption{Camera layouts and sample views of the datasets used in this work.}
\end{figure*}

CityFlow \cite{tang2019cityflow} is one of the earliest and most representative 
open-source MCVT datasets, released by NVIDIA Research in 2019. Its test-set layout is shown in~\cref{fig:cityflow}. The dataset contains 
approximately 3.25 hours of 960p traffic video recorded from 46 cameras across 
10 intersections in a medium-sized U.S.\ city, covering roughly 2.5\,km of roadway. 
In our experiment, we use the training and validation splits (40 cameras) to 
fine-tune the feature extractor, and the six-camera test split to evaluate the 
performance of the proposed EASE-MCVT pipeline.

\subsection{Implementation Details}
\subsubsection{Hardware settings}
Due to hardware availability constraints, the EASE-MCVT framework is evaluated in 
a laboratory environment using a single NVIDIA Jetson AGX Orin 64\,GB module as 
the edge computing device and a workstation equipped with an Intel 13th Gen 
i7-13700KF CPU and an NVIDIA GeForce RTX~4070\,Ti GPU as the central server, 
running Ubuntu~20.04.4~LTS. The edge node and the central server are connected 
via a wired Ethernet link.

In this paper, we emulate a multi-camera deployment by running the edge pipeline per camera, recording the resulting message streams, and replaying multiple streams concurrently to simulate multi-node operation. In parallel, the framework is also being tested for real-world deployment in several European cities.

\subsubsection{Hyperparameter settings}
Unless otherwise stated, the hyperparameter settings in this section were selected empirically through preliminary experiments, with the aim of balancing accuracy, runtime stability, and practical deployment constraints.
\begin{itemize}
    \item \textbf{Object Detection.}  
    We follow the standard YOLO inference pipeline. The confidence threshold is 
    set to 0.35 and the IoU threshold for non-maximum suppression is 0.40. The 
    image inference size is configured as $1120 \times 672$ to improve detection 
    of small, distant vehicles. All remaining settings follow the default YOLO 
    configuration. The motion detector threshold $\tau_{\mathrm{fg}}$ is set to 
    1e-4. 

    \item \textbf{Single-Camera Tracking.}  
    We adopt the default parameter settings from BoxMOT \cite{brostrom2025boxmot} 
    for the single-camera tracking module.

    \item \textbf{Geo-Mapping.}  
    The vehicle center-of-gravity height is assumed to be 0.5 meters.

    \item \textbf{Feature Extraction.}  
    For the dynamic workload adjustment scheme, ${Q_{\max}(\Delta)}$ is set to 25 images within an allowed processing window of 1 second.

    \item \textbf{Single-Camera Rematch.}  
    For refining fragmented single-camera tracklets, we use a temporal threshold 
    of $T_{\mathrm{th}} = 3$\,s (45 frames), a spatial threshold of 
    $D_{\mathrm{th}} = 0.2$ of the image width, and a feature-similarity 
    threshold of $F_{\mathrm{th}} = 0.2$.

    \item \textbf{Self-Supervised Camera Link Model.}  
    The bandwidth parameter for the KDE-based transition-time estimation is set to 
    5.0.

    \item \textbf{Cross-Camera Association.}  
    The association process is executed every $\alpha = 2500$\ frames. The cost-function
    weights are set to $\delta = 1$ and $\epsilon = 1$, and the greedy-matching 
    distance threshold is 0.4.
\end{itemize}
\subsection{EASE-MCVT Framework Performance Evaluation}

\subsubsection{Edge-Side Performance}

\paragraph{Object Detection}
We evaluate the edge-side vehicle detection module on RoundaboutHD in terms of detection accuracy and runtime. All results are obtained using our fine-tuned YOLOv11-nano detector deployed with TensorRT (FP16 quantization) on an NVIDIA Jetson AGX Orin, with an input image size of $1120\times672$. Detection accuracy is measured using mean average precision (mAP) averaged over IoU thresholds from 0.5 to 0.95. Runtime is reported as the average per-frame inference latency (ms). We compare our detector against the non-fine-tuned YOLOv11X baseline provided with RoundaboutHD; results are summarized in~\cref{tab:det-eval}.

\begin{table}[h]
\centering
\caption{Edge-side object detection evaluation on RoundaboutHD. $\uparrow$ indicates that higher values are better, and $\downarrow$ indicates that lower values are better.}
\label{tab:det-eval}
\resizebox{\linewidth}{!}{%
\begin{tabular}{lcccccccc}
\toprule
\textbf{Metric} 
& \multicolumn{2}{c}{\textbf{Cam01}} 
& \multicolumn{2}{c}{\textbf{Cam02}} 
& \multicolumn{2}{c}{\textbf{Cam03}} 
& \multicolumn{2}{c}{\textbf{Cam04}} \\
\cmidrule(lr){2-3}\cmidrule(lr){4-5}\cmidrule(lr){6-7}\cmidrule(lr){8-9}
& \textbf{Baseline} & \textbf{Ours}
& \textbf{Baseline} & \textbf{Ours}
& \textbf{Baseline} & \textbf{Ours}
& \textbf{Baseline} & \textbf{Ours} \\
\midrule
mAP@0.5:0.95 $\uparrow$
& 42.08 & 56.47
& 70.39 & 66.62
& 74.39 & 65.44
& 76.60 & 51.25 \\
Latency (ms/frame) $\downarrow$
& \textit{Offline} & 25.84
& \textit{Offline} & 25.68
& \textit{Offline} & 25.50
& \textit{Offline} & 25.68 \\
\bottomrule
\end{tabular}}
\end{table}

Overall, our implementation achieves low inference latency on the edge device while maintaining good detection accuracy for real-time deployment. We do not report motion-detector results on RoundaboutHD because the traffic flow in the videos is continuous and rarely contains idle periods, so frame skipping would provide little benefit.

\paragraph{Single-Camera Tracking}
We evaluate the single-camera tracking module on RoundaboutHD by comparing our online, edge-deployed tracker against the offline SCT baseline reported with the dataset. We use ByteTrack as the default tracker in EASE-MCVT. Tracking accuracy is measured using IDF1 \cite{dukemtmct}, and runtime is reported as the average per-frame tracking latency (ms) measured on the Jetson AGX Orin. Results for each camera are summarized in~\cref{tab:sct-eval}.

\begin{table}[h]
\centering
\caption{Edge-side single-camera tracking evaluation on RoundaboutHD. $\uparrow$ indicates that higher values are better, and $\downarrow$ indicates that lower values are better.}
\label{tab:sct-eval}
\resizebox{\linewidth}{!}{%
\begin{tabular}{lcccccccc}
\toprule
\textbf{Metric} 
& \multicolumn{2}{c}{\textbf{Cam01}} 
& \multicolumn{2}{c}{\textbf{Cam02}} 
& \multicolumn{2}{c}{\textbf{Cam03}} 
& \multicolumn{2}{c}{\textbf{Cam04}} \\
\cmidrule(lr){2-3}\cmidrule(lr){4-5}\cmidrule(lr){6-7}\cmidrule(lr){8-9}
& \textbf{Baseline} & \textbf{Ours}
& \textbf{Baseline} & \textbf{Ours}
& \textbf{Baseline} & \textbf{Ours}
& \textbf{Baseline} & \textbf{Ours} \\
\midrule
IDF1$\uparrow$
& 72.60 & 64.10 
& 90.50 & 66.30 
& 84.20 & 55.80 
& 86.00 & 73.40 \\
Latency (ms/frame)$\downarrow$
& \textit{Offline} & 5.48
& \textit{Offline} & 5.66
& \textit{Offline} & 5.80
& \textit{Offline} & 3.23 \\
\bottomrule
\end{tabular}}
\end{table}

As expected, the offline baseline achieves higher IDF1, as it can use heavier models and non-real-time processing. In contrast, our edge implementation delivers low per-frame latency (3--6\,ms) while maintaining reasonable identity consistency, providing a practical SCT component for real-time deployment.

\paragraph{Feature Extraction}
We evaluate the feature extraction module on RoundaboutHD in terms of latency and runtime stability. Two latency metrics are reported. \textit{Frame-level latency} is defined as the elapsed time between the arrival of a frame-level SAE message at the \textit{feature extractor} and the completion of its processing. \textit{Tracklet-level latency} measures the elapsed time between the arrival of a frame-level message and the transmission of the corresponding completed tracklet information, averaged over all frames within the tracklet. A tracklet is considered complete when no new detection has been associated with it for a fixed timeout of 3000\,ms. The results are reported in \cref{tab:feature-e2e}.

\begin{table}[h] 
\centering 
\caption{Feature extraction latency for the four camera streams.} 
\label{tab:feature-e2e} 
\begin{tabular}{lcccc} 
\toprule 
\textbf{Latency (ms) $\downarrow$} & \textbf{Cam01} & \textbf{Cam02} & \textbf{Cam03} & \textbf{Cam04} \\ 
\midrule 
Frame-level & 121.34 & 93.28 & 111.59 & 67.95 \\ 
Tracklet-level & 13567.59 & 11219.26 & 10571.93 & 6691.52 \\ 
\bottomrule 
\end{tabular} 
\end{table}

In addition to direct latency measurements, we evaluate runtime stability by monitoring the FIFO input queue of the feature extraction module. Queue occupancy indicates whether the module can keep pace with incoming tracklet-update messages under the proposed \textit{dynamic workload scheme}. The average and maximum queue occupancies are reported in \cref{tab:fx-eval}.

\begin{table}[h]
    \centering
    \caption{Edge-side feature extraction queue occupancy on RoundaboutHD. The queue capacity is 64 messages.}
    \label{tab:fx-eval}
    \begin{tabular}{lcccc}
        \toprule
        \textbf{Metric (messages)} & \textbf{Cam01} & \textbf{Cam02} & \textbf{Cam03} & \textbf{Cam04} \\
        \midrule
        Avg. queue occupancy$\downarrow$ & 2.25 & 2.25 & 2.20 & 1.39 \\
        Max. queue occupancy$\downarrow$ & 22 & 18 & 16 & 21 \\
        \bottomrule
    \end{tabular}
\end{table}

Across the four cameras, the average frame-level latency is 98.54\,ms. Although this is greater than the 66.67\,ms frame interval of a 15-FPS stream, the queue remains bounded and no overflow is observed, indicating that the feature extraction module maintains sufficient throughput to continuously process the incoming workload. The average tracklet-level latency is 10512.58\,ms. This larger delay mainly results from the tracklet-based processing design, as completed tracklet information is only transmitted after all observations have been received and the fixed 3000\,ms termination timeout has expired.

\paragraph{Edge-Side Latency Evaluation}
We further evaluate the end-to-end latency of the complete edge pipeline. Similar to the feature extraction evaluation, we report \textit{frame-level pipeline latency} and \textit{tracklet-level pipeline latency}. The former measures the elapsed time between a frame entering the \textit{video-source} module and the completion of its frame-level processing at the \textit{feature extractor}, including both computation and inter-module communication. The latter measures the delay between frame arrival and the transmission of its corresponding completed tracklet representation. The results are shown in \cref{tab:edge-e2e}.

\begin{table}[h] 
\centering 
\caption{Edge-side E2E latency for the four camera streams.} 
\label{tab:edge-e2e} 
\begin{tabular}{lcccc} 
\toprule 
\textbf{E2E latency (ms) $\downarrow$} & \textbf{Cam01} & \textbf{Cam02} & \textbf{Cam03} & \textbf{Cam04} \\ 
\midrule 
Frame-level & 299.25 & 266.11 & 292.92 & 240.96 \\ 
Tracklet-level & 13880.11 & 11481.60 & 10862.85 & 7236.64 \\ 
\bottomrule 
\end{tabular} 
\end{table}

The average frame-level end-to-end latency is 274.81\,ms, corresponding to approximately 4.12 frame intervals at 15 FPS. This value represents pipeline delay rather than processing throughput. Combined with the bounded feature-extraction queue and zero overflow events, the results indicate that the edge pipeline can continuously sustain the evaluated 15-FPS workload while introducing a sub-second processing delay. The average tracklet-level latency is 10865.30\,ms, which is mainly determined by the tracklet duration and the fixed 3000\,ms termination timeout. Therefore, the tracklet-level metric primarily characterizes the delay before a completed vehicle representation becomes available for server-side association.

\paragraph{Bandwidth Evaluation}
In an edge--server architecture, bandwidth is a key scalability constraint: higher transmission rates increase network cost and can amplify queuing delay or message loss under high data-volume conditions. To quantify the bandwidth savings of EASE-MCVT, we compare two edge-to-server payload settings: (i) embedding-based tracklet metadata, which includes detection results, frame-level embeddings, and tracklet-level embeddings, and (ii) JPEG-based tracklet metadata, which includes tracklet information and JPEG images. We measure the total transmitted payload over a 10-minute window and report the corresponding average throughput in MB/s. Results are summarized in Table~\ref{tab:bandwidth}.

\begin{table}[h]
    \centering
    \caption{Edge-to-server bandwidth on RoundaboutHD.}
    \label{tab:bandwidth}
    \resizebox{\linewidth}{!}{%
    \begin{tabular}{lcccc}
        \toprule
        \textbf{Setting (MB/s)$\downarrow$} & \textbf{Cam01} & \textbf{Cam02} & \textbf{Cam03} & \textbf{Cam04} \\
        \midrule
        Embedding-Based Tracklet Metadata 
            & 0.26 & 0.19 & 0.25 & 0.25 \\
        JPEG-Based Tracklet Metadata 
            & 10.35 & 9.12 & 11.57 & 9.18 \\
        \bottomrule
    \end{tabular}}
\end{table}

Overall, transmitting embeddings instead of frame-level JPEGs reduces the required bandwidth by approximately $42\times$ (averaged across cameras). This reduction directly improves scalability by lowering per-camera bandwidth requirements, which makes it easier to support larger camera networks under practical network constraints.

\subsubsection{Central Server Performance}
\begin{figure}[h]
    \centering
    \includegraphics[width=1\linewidth]{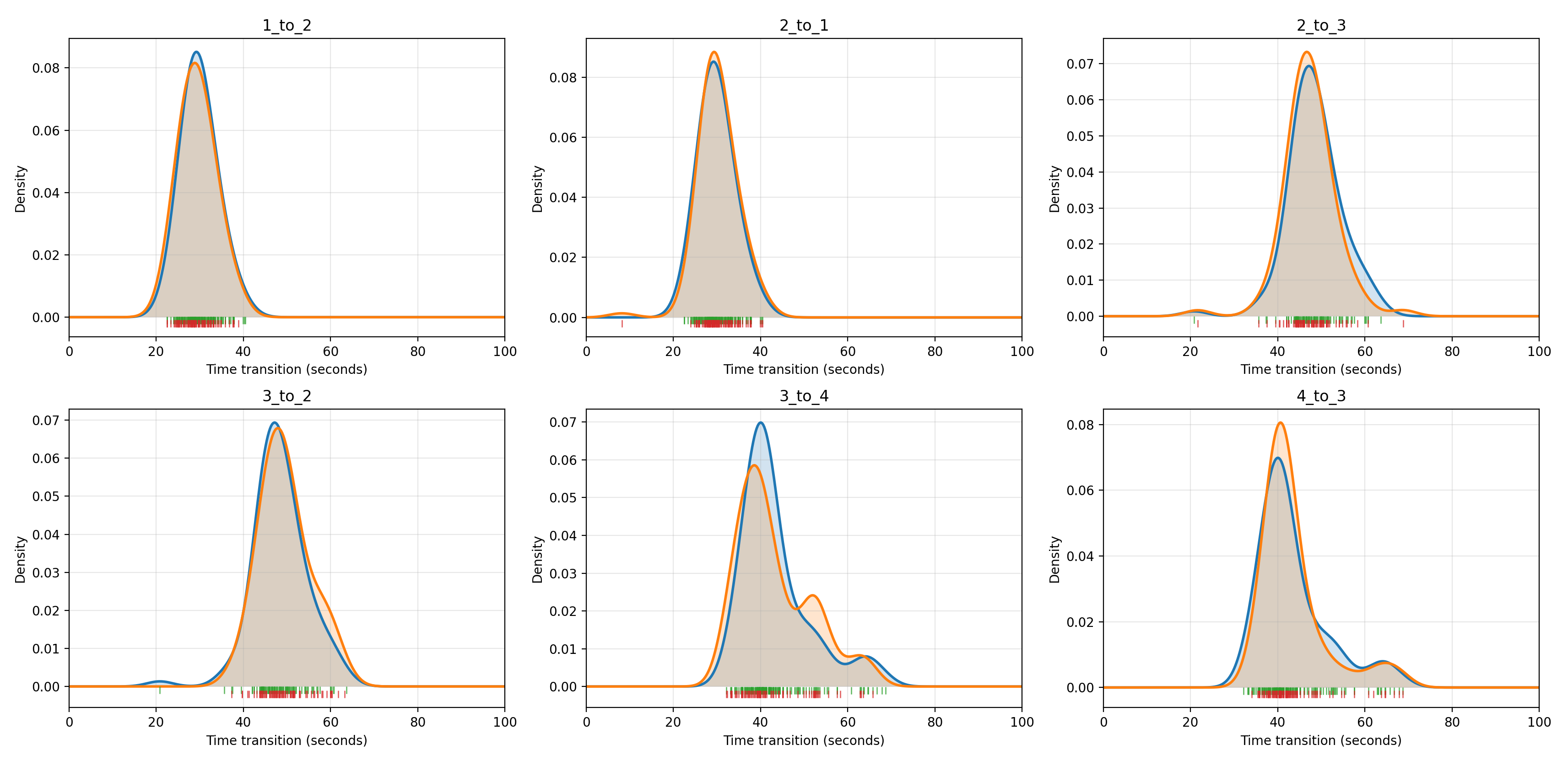}
    \caption{Ground-truth versus estimated transition-time distributions for the self-supervised camera link model. The orange region shows the ground-truth transition-time density, and the blue region shows the estimated KDE from the proposed camera link model. The green vertical bars indicate the estimated transition times, while the red vertical bars indicate the ground-truth transition times.}
    \label{fig:CLM_res}
\end{figure}

\paragraph{Self-Supervised Camera Link Model}
To provide a qualitative evaluation for the self-supervised camera link model, we compare the estimated transition-time density with a ground-truth transition-time density computed from the RoundaboutHD annotations. Specifically, for each adjacent camera link, we derive ground-truth transition times from the RoundaboutHD annotations; we then overlay this ground-truth density with the kernel density estimate produced by our camera link model. The qualitative comparison is shown in~\cref{fig:CLM_res}.

As shown in~\cref{fig:CLM_res}, the estimated transition-time densities closely match the ground-truth distributions across links, indicating that the proposed camera link model captures the underlying spatial--temporal relationship between adjacent cameras and provides reliable temporal constraints for downstream cross-camera association.

\paragraph{Multi-Camera Tracking}
We evaluate the accuracy of EASE-MCVT on both \textit{RoundaboutHD} and \textit{CityFlow}. As shown in~\cref{tab:mcvt-comparison}, we compare EASE-MCVT with several representative offline state-of-the-art MCVT methods. In addition, to quantify the contribution of our server-side add-on modules, we conduct an ablation study on the \textit{single-camera re-match} module and the \textit{self-supervised camera link model (CLM)}, with results reported in~\cref{tab:mcmt-ablation}.

\begin{table}[!t]
    \centering
    \caption{Comparison of multi-camera vehicle tracking performance on RoundaboutHD and CityFlow.}
    \label{tab:mcvt-comparison}
    \resizebox{\linewidth}{!}{%
    \begin{tabular}{llcccc}
        \toprule
        \textbf{Dataset} & \textbf{Method} & \textbf{IDF1$\uparrow$} & \textbf{IDP$\uparrow$} & \textbf{IDR$\uparrow$}  \\
        \midrule
        \multirow{4}{*}{RoundaboutHD}
        & ELECTRICITY~\cite{qian2020electricity} & 28.14 & 26.45 & 30.06  \\
        & AIC21-MCVT~\cite{liu2021citywin}   & 36.92 & 36.54 & 37.30  \\
        & AIC22-MCVT~\cite{li2022multi}   & 45.19 & 37.00 & 58.04 \\
        & \textbf{EASE-MCVT (Ours)}    & \textbf{71.63} & \textbf{80.60} & \textbf{64.44} \\
        \midrule
        \multirow{4}{*}{CityFlow}
        & ELECTRICITY~\cite{qian2020electricity} & 45.85 & -     & -     \\
        & AIC21-MCVT~\cite{liu2021citywin}   & 80.95 & 85.69 & 76.70 \\
        & AIC22-MCVT~\cite{li2022multi}    & \textbf{84.37} & \textbf{89.00} & \textbf{80.20} \\
        & EASE-MCVT (Ours)    & 55.17 & 75.99 & 43.31 \\
        \bottomrule
    \end{tabular}}
\end{table}

\begin{table}[!t]
    \centering
    \caption{Ablation study of add-on modules on RoundaboutHD and CityFlow.}
    \label{tab:mcmt-ablation}
    \resizebox{\linewidth}{!}{%
    \begin{tabular}{lcccccc}
        \toprule
        \textbf{Dataset} & \textbf{Baseline} & \thead{\textbf{+SCT}\\\textbf{Re-Match}} & \textbf{+CLM} & \textbf{IDF1$\uparrow$} & \textbf{IDP$\uparrow$} & \textbf{IDR$\uparrow$} \\
        \midrule
        \multirow{4}{*}{RoundaboutHD}
        & \checkmark & - & - & 64.92 & 74.79 & 57.34 \\
        & \checkmark & \checkmark & - & 68.75 (+3.83) & 77.42 (+2.63) & 61.84 (+4.50) \\
        & \checkmark & - & \checkmark & 66.64 (+1.72) & 77.75 (+2.96) & 58.30 (+0.96) \\
        & \checkmark & \checkmark & \checkmark & 71.63 (+6.71) & 80.60 (+5.81) & 64.44 (+7.10) \\
        \midrule
        \multirow{4}{*}{CityFlow}
        & \checkmark & - & - & 43.66 & 40.72 & 47.06 \\
        & \checkmark & \checkmark & - & 43.66 (+0.00) & 41.03 (+0.31) & 46.47 (-0.59) \\
        & \checkmark & - & \checkmark & 54.51 (+10.85) & 75.40 (+34.68) & 42.68 (-4.38) \\
        & \checkmark & \checkmark & \checkmark & 55.17 (+11.51) & 75.99 (+35.27) & 43.31 (-3.75) \\
        \bottomrule
    \end{tabular}}
\end{table}

For the overall MCVT comparison in~\cref{tab:mcvt-comparison}, EASE-MCVT achieves the best performance on \textit{RoundaboutHD}, reaching an IDF1 score of 71.63\%. This result is consistent with our design goal, indicating that the proposed combination of dataset-adapted edge-side modules and server-side association refinements can achieve strong tracking accuracy on this real-world roundabout dataset without sacrificing deployment practicality. On \textit{CityFlow}, EASE-MCVT achieves 55.17\% IDF1, which is higher than \textit{ELECTRICITY} but lower than \textit{AIC21-MCVT} and \textit{AIC22-MCVT}. We note that the compared methods were developed as offline systems specifically for the AI City Challenge setting on \textit{CityFlow}, whereas EASE-MCVT is designed primarily for real-time and scalable deployment rather than maximizing offline benchmark performance. From this perspective, although EASE-MCVT does not match the best offline results on \textit{CityFlow}, its performance still demonstrates reasonable generalizability across different camera-network settings, while preserving the practical advantages of a deployment-oriented framework.

% Beyond tracking accuracy, our framework also demonstrates stable runtime behavior during online operation. Following the same queue-based evaluation methodology used for the feature-extraction module, we monitor the server-side input queue occupancy during multi-camera tracking. On \textit{RoundaboutHD}, the maximum queue occupancy is 12 and the average queue occupancy is 0.21. On \textit{CityFlow}, the maximum queue occupancy increases to 53 and the average queue occupancy to 1.39, reflecting the heavier workload introduced by the larger camera network and denser traffic. Importantly, we observe zero queue overflow events in both settings. This indicates that, under the tested conditions, the central server can process incoming tracklet updates at least as fast as they arrive, providing evidence that EASE-MCVT supports real-time throughput in the sense of stable end-to-end operation. In contrast, the compared baseline methods are designed for offline processing and therefore do not target this deployment setting.

The ablation results in~\cref{tab:mcmt-ablation} further verify the effectiveness of the proposed server-side modules. On \textit{RoundaboutHD}, both add-on modules improve performance over the baseline, and the combination of \textit{SCT Re-Match} and \textit{CLM} yields the best result, improving IDF1 from 64.92\% to 71.63\%. This indicates that reconnecting fragmented single-camera tracklets and introducing learned spatial--temporal constraints are both beneficial for downstream cross-camera association. On \textit{CityFlow}, the overall trend is similar: adding \textit{CLM} produces a substantial gain, increasing IDF1 from 43.66\% to 54.51\%, and combining both modules further improves IDF1 to 55.17\%. In contrast, the \textit{SCT Re-Match} module alone brings only a marginal improvement on this dataset. A likely reason is that the edge-side SCT outputs on \textit{CityFlow} are less reliable due to the more challenging visual conditions, including lower video quality, denser traffic, and more frequent occlusions, which limit the amount of recoverable fragmentation through conservative same-camera re-linking. Overall, the ablation study demonstrates that both add-on modules contribute positively to the proposed framework and validates the effectiveness of their designs in improving multi-camera tracking performance.

\paragraph{Server-Side Scalability and Latency Evaluation}
We evaluate the scalability of the central-server pipeline by increasing the number of camera streams from the original four-camera RoundaboutHD configuration to 8, 16, and 32 cameras. To construct these larger settings, we duplicate the edge-side messages two, four, and eight times, respectively. The additional cameras are configured as consecutive adjacent camera pairs, extending the original camera topology. This setup increases both the incoming message volume and the number of candidate cross-camera associations while preserving the temporal and spatial characteristics of the original data.

\begin{figure}[h]
    \centering
    \includegraphics[width=1\linewidth]{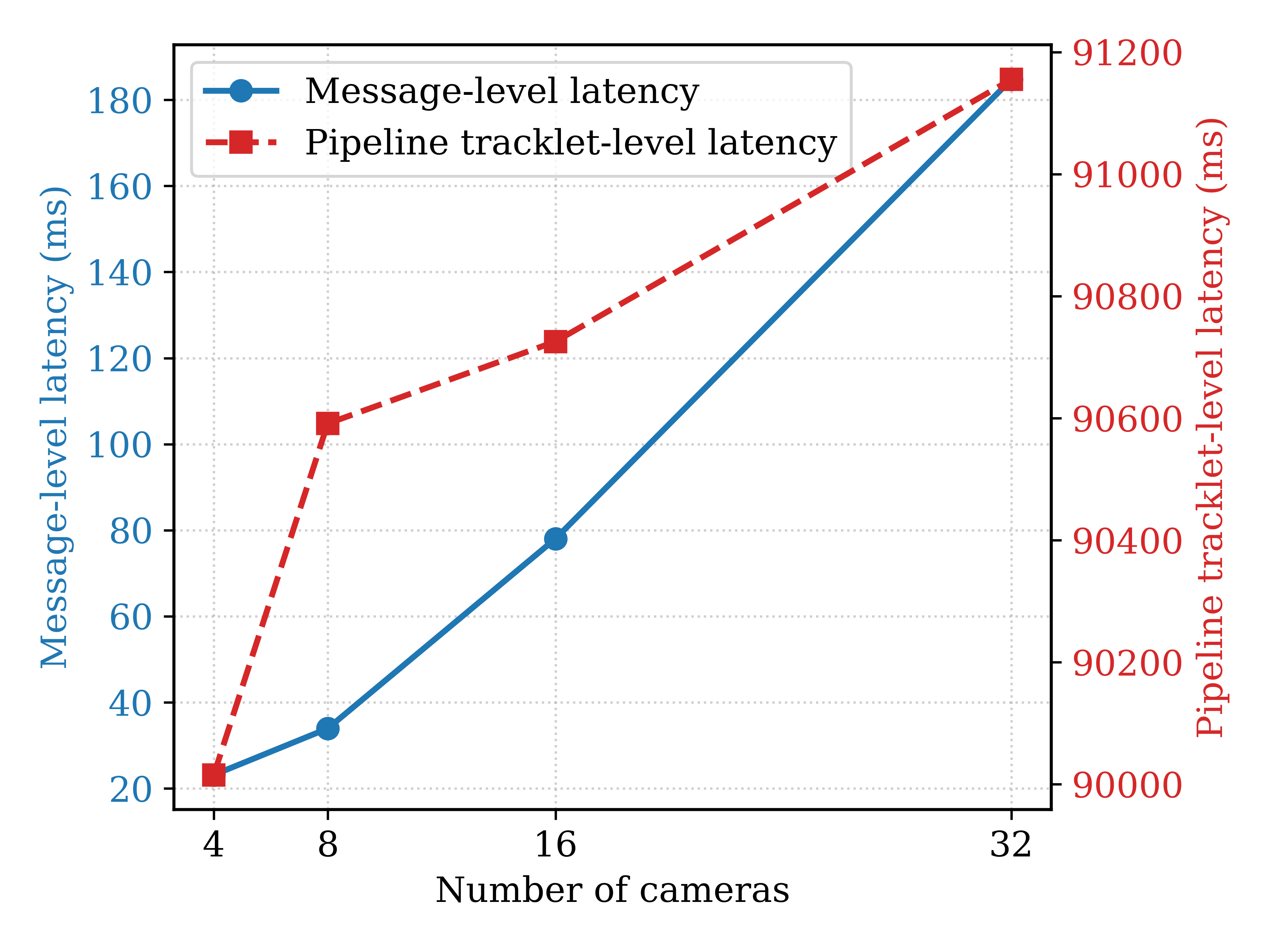}
    \caption{Central-server latency under different numbers of camera streams. The left y-axis shows the message-level latency, and the right y-axis shows the tracklet-level latency.}
    \label{fig:scalability}
\end{figure}

We report two latency metrics. \textit{Message-level latency} is defined as the elapsed time between the arrival of an SAE message from an edge node and the completion of its processing by the central server. \textit{Tracklet-level latency} is defined as the elapsed time between the arrival of an SAE message and the completion of the corresponding cross-camera association round, averaged over all tracklets. A tracklet is considered processed once it has been evaluated in an association round, regardless of whether a valid cross-camera match is identified. The results are shown in \cref{fig:scalability}.

To further evaluate runtime stability, we monitor the server-side input queue under each scalability setting. The average and maximum queue occupancies are reported in \cref{tab:server-queue}.

\begin{table}[h]
\centering
\caption{Server-side queue occupancy under different numbers of camera streams.}
\label{tab:server-queue}
\begin{tabular}{lcccc}
\toprule
\textbf{Metric} & \textbf{4 Cam} & \textbf{8 Cam} & \textbf{16 Cam} & \textbf{32 Cam} \\
\midrule
Avg. queue occupancy $\downarrow$ & 0.013 & 0.034 & 0.126 & 0.160 \\
Max. queue occupancy $\downarrow$ & 4 & 8 & 28 & 30 \\
\bottomrule
\end{tabular}
\end{table}

As shown in \cref{fig:scalability}, the message-level latency increases with the number of cameras as the server receives more messages and performs more candidate cross-camera associations. In the 32-camera setting, the average message-level latency is approximately 180\,ms. Despite the increased workload, the input queue remains bounded across all evaluated settings, with an average occupancy of only 0.16 messages and a maximum occupancy of 30 messages in the 32-camera setting. Together, these results indicate that the central server can sustain the evaluated workload with stable processing throughput.

The tracklet-level latency remains approximately 90\,s across the evaluated settings. This substantially larger latency is mainly determined by the periodic association strategy rather than computational processing. Cross-camera association is triggered every $\alpha=2500$ frames, corresponding to an interval of approximately 166.7\,s at 15 FPS. Assuming tracklet arrivals are approximately uniformly distributed within an association interval, the expected waiting time before the next association round is approximately half of this interval, or 83.3\,s (1.39 minutes). Therefore, most of the observed tracklet-level latency is introduced by the application-specific association interval, while the remaining delay is attributable to message processing and cross-camera association.

Overall, the results demonstrate that the proposed server-side pipeline maintains stable real-time throughput under the evaluated workloads, supporting up to 32 emulated camera streams on the tested consumer-grade workstation.

\section{Conclusion}
This work presents EASE-MCVT, an edge--server multi-camera vehicle tracking framework designed for real-time and scalable ITS deployment. EASE-MCVT combines algorithmic and system-level adaptations across the edge and server sides. On the edge side, lightweight detection and tracking are integrated with tracklet-level feature extraction and adaptive workload control to reduce computational overhead. On the server side, single-camera re-matching and a self-supervised camera link model improve the robustness of cross-camera association, while the overall framework is implemented within the SAE-ENGINE architecture to support standardized communication, deployment, and data management. Experimental results demonstrate that EASE-MCVT achieves a practical balance between tracking accuracy and deployment efficiency. The framework reaches 71.63\% IDF1 on \textit{RoundaboutHD}, outperforming the compared methods, and achieves 55.17\% IDF1 on \textit{CityFlow}. On the edge device, the complete pipeline sustains the evaluated 15-FPS workload with an average frame-level latency of 274.81\,ms while maintaining bounded feature-extraction queues. On the server side, the framework remains stable under workloads of up to 32 emulated camera streams, with an average message-level latency of approximately 180\,ms at the largest evaluated scale and bounded queue occupancy. Overall, these results demonstrate that real-time MCVT can be achieved through coordinated optimization of both algorithmic processing and distributed system design, rather than by tracking accuracy alone. EASE-MCVT therefore provides a deployment-oriented foundation for scalable multi-camera traffic monitoring. Future work will further validate the framework through long-term, large-scale physical camera deployments and investigate adaptive resource allocation under more heterogeneous traffic and hardware conditions.
\section*{Acknowledgment}
Yuqiang Lin and Sam Lockyer are supported by a scholarship from the EPSRC Centre for Doctoral Training in Advanced Automotive Propulsion Systems (AAPS) under project EP/S023364/1.

\bibliographystyle{IEEEtran}
\bibliography{example}

\end{document}